\documentclass[twocolumn]{article}
\usepackage{spconf,amsmath,graphicx}
\usepackage{afterpage}

\usepackage{caption}
\usepackage{xurl}

\usepackage{fancyhdr}
 
\fancypagestyle{firstpagefooter}{
  \fancyhf{}  
  \fancyfoot[L]{\scriptsize
    © 2025 IEEE. Personal use of this material is permitted. 
    Permission from IEEE must be obtained for all other uses, in any current or future media, 
    including reprinting/republishing this material for advertising or promotional purposes, 
    creating new collective works, for resale or redistribution to servers or lists, 
    or reuse of any copyrighted component of this work in other works.
  }

}

\title{TARGET DRIVEN ADAPTIVE LOSS FOR INFRARED SMALL TARGET DETECTION}
\name{Yuho Shoji, Takahiro Toizumi, and Atsushi Ito}
\address{NEC Corporation}
\begin{document}
\maketitle
\thispagestyle{firstpagefooter}

\begin{abstract}
We propose a target driven adaptive (TDA) loss to enhance the performance of infrared small target detection (IRSTD). Prior works have used loss functions, such as binary cross-entropy loss and IoU loss, to train segmentation models for IRSTD. Minimizing these loss functions guides models to extract pixel-level features or global image context. However, they have two issues: improving detection performance for local regions around the targets and enhancing robustness to small scale and low local contrast. To address these issues, the proposed TDA loss introduces a patch-based mechanism, and an adaptive adjustment strategy to scale and local contrast. The proposed TDA loss leads the model to focus on local regions around the targets and pay particular attention to targets with smaller scales and lower local contrast. We evaluate the proposed method on three datasets for IRSTD. The results demonstrate that the proposed TDA loss achieves better detection performance than existing losses on these datasets.

\end{abstract}
\section{Introduction}
\label{sec:intro}
Small target detection is crucial in various applications, including maritime rescue and traffic management \cite{zhao2022single, zhang2021review}. Especially, infrared small target detection (IRSTD) has gained attention because of its robustness against varying light conditions. IRSTD focuses on identifying targets typically occupying less than 50 pixels in an image, a challenge arising from long-distance infrared imaging. These small targets often lack distinctive features and are easily obscured by complex backgrounds. Detection becomes more difficult due to two factors: the smaller scale and the low local contrast between targets and their immediate surroundings \cite{yu2022pay, dai2021attentional, hou2021ristdnet}.

Conventional methods to IRSTD can be broadly categorized into filtering-based, local contrast-based, and low-rank-based methods \cite{zhang2021review,  han2019local, dai2017reweighted}. While these methods have shown some success, they often rely on hand-crafted features and have difficulty generalizing across diverse scenarios. Recently, deep learning-based methods have gained prominence in IRSTD thanks to their ability to automatically learn robust features and generalize well to various environments \cite{zhang2022isnet, liu2024infrared, li2022dense, wu2022uiu, wang2019miss}. We focus on deep learning-based methods because of their extensibility and potential for improving robustness.

Current deep learning-based methods for IRSTD have mainly focused on designing sophisticated network architectures to extract more discriminative features. For example, some prior research has introduced attention mechanisms, dense-nested structures, multi-scale heads, and spatial-channel cross transformers \cite{yu2022pay, dai2021attentional, li2022dense, yuan2024sctransnet}. While loss functions are also important for extracting discriminative features, fewer studies have focused on this aspect, indicating a potential for improvement. Studies on loss functions can enhance detection performance without increasing architectural complexity. This offers a complementary approach to architecture-based methods.

Existing losses for IRSTD can be broadly classified into pixel level metric losses (e.g., binary cross-entropy loss) and global image level metric losses (e.g., IoU loss and Dice loss) \cite{liu2024infrared, wu2022uiu, dai2021asymmetric}. While these losses enable a detection model to capture pixel-level features and global image context, they have two issues. Firstly, the models trained by these losses often misclassify the edge regions of targets and the background areas immediately surrounding the targets. These areas are often mistaken due to their close spatial proximity. Furthermore, because the pixels of these areas are relatively few, they are not dominant in conventional losses and are often overlooked. Secondly, models trained with existing losses frequently misclassify targets that have small scale or low local contrast. Small-scale targets occupy fewer pixels and thus contribute less to the overall loss, causing the model to focus on them insufficiently. Targets with low local contrast are inherently harder to distinguish from their cluttered background. Moreover, deep networks are empirically known to learn easy patterns first and then postpone harder ones \cite{arpit2017closer}. Consequently, these hard targets receive limited updates during training, and the detector fails in tough cases, resulting in degraded performance.

To solve these problems, we propose a target driven adaptive (TDA) loss designed to improve detection performance in local regions and detection rates for targets with small scales and low local contrasts. The proposed TDA loss employs a patch-based mechanism and minimizing this loss leads the model to effectively separate targets from their surrounding backgrounds. Furthermore, we introduce a dynamic loss strategy, which adapts to the scale and local contrast of each target. This dynamic loss ensures that the model pays particular attention to targets with small scales and low local contrast. The proposed TDA loss can be used with existing global image metric losses and combining them enhances detection rate for challenging cases and improves detection performance for overall images. We evaluate the proposed TDA loss against existing losses using three datasets: IRSTD-1k, SIRST-v1, and SIRST-v2 \cite{ dai2021asymmetric, wang2022iaanet, dai2023one}. The experimental results demonstrate that our approach improves detection performance for targets with small scales and low local contrast. Moreover, it maintains or improves the detection performance across all types of targets.

\section{Related work}
Loss functions are crucial for improving detection performance. Some prior works have introduced loss functions, such as the adversarial loss, the edge loss, and the object-detection-based loss \cite{zhang2022isnet, wang2019miss, wang2022iaanet}. These losses are developed for specific network architectures. Applying these losses to different network designs is challenging, which restricts their versatility. 
We focus on loss functions applicable to general segmentation models with identical input and output sizes, designed for IRSTD. 
We divide existing losses into global image level and pixel level losses.

Global image level metric losses optimize metrics defined for the entire image. Prominent examples include IoU and Dice losses, which optimize intersection over union and Dice coefficient, respectively \cite{huang2019batching, sudre2017generalised}. Tversky loss generalizes Dice loss by reweighting false negatives and false positives to mitigate the impact of class imbalance \cite{lin2017focal}. The scale and location sensitive loss (SLS loss) reduce discrepancies in scale and location between predicted and actual target regions \cite{liu2024infrared}. The SLS loss achieved better detection performance than IoU loss and Dice loss. These losses aim to distinguish small targets from the overall background. However, global image level metric losses have limitations in addressing multi-target scenarios. In such scenarios, smaller targets contribute less to the overall loss and are often neglected during training.

Pixel level metric losses are the image wide average of losses defined at each pixel. While this approach facilitates learning detailed features, it faces challenges in the IRSTD due to significant class imbalance. For instance, models trained with the binary cross-entropy (BCE) loss often fail to improve the detection performance because they develop a bias towards classifying all pixels as background. The Focal loss attempts to address this issue by adjusting sample weights, emphasizing misclassified samples over correctly classified samples \cite{lin2017focal}.

Prior research has shown that the ability to learn local contextual information is crucial for enhancing detection performance \cite{yu2022pay, dai2021attentional, zhang2022isnet}. Existing loss functions do not pay particular attention to local regions around targets. We propose a loss that specifically focus on local regions around targets.

Scale-adaptive IoU (SIoU) loss has been proposed to improve the detection performance for small objects in object detection tasks \cite{jeune2023rethinking}. The SIoU loss value increases for objects with smaller bounding boxes, leading the model to focus more on smaller objects. Inspired by the SIoU loss, we design an adaptive loss for segmentation tasks that focuses on targets with smaller scales and lower local contrasts.

\begin{figure}[t]
  \begin{center}
   \includegraphics[width=0.99\linewidth]{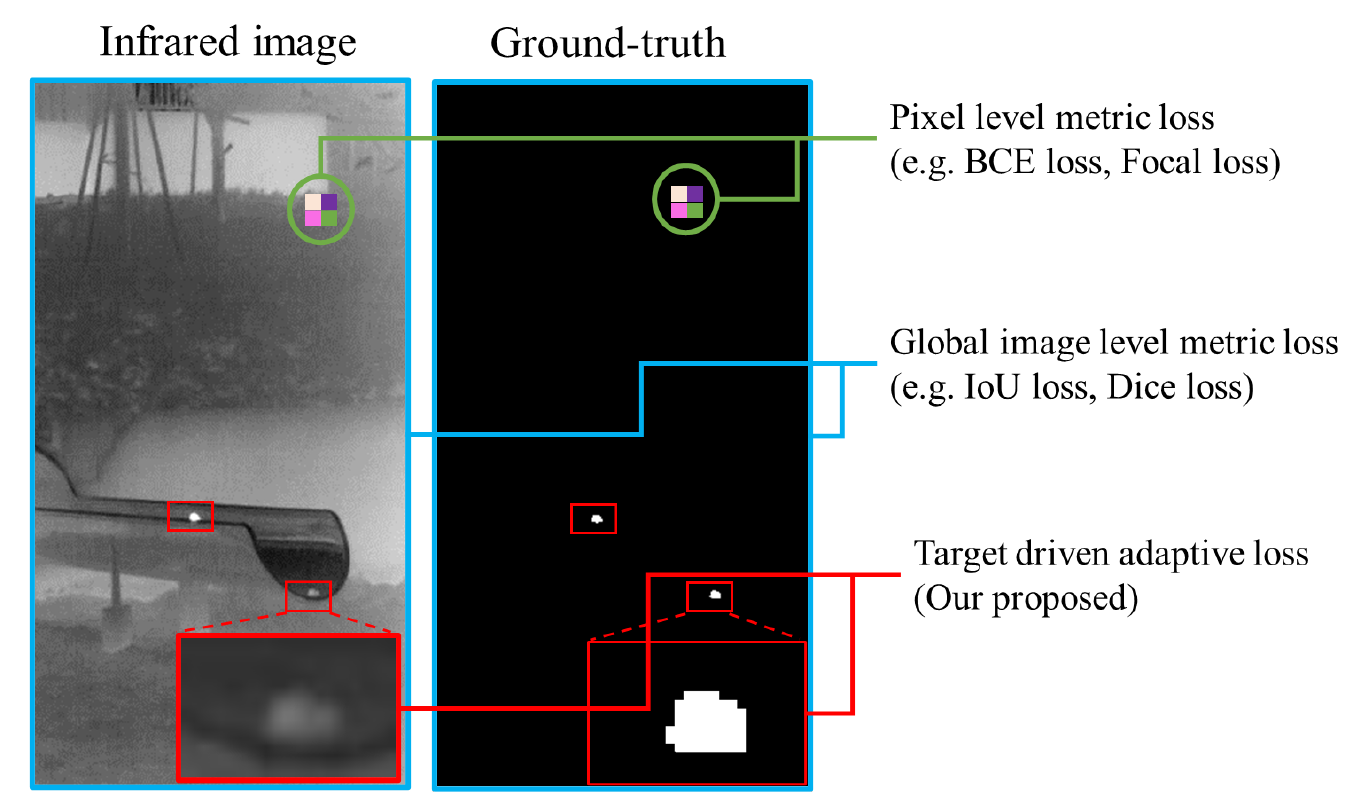}
  \end{center}
  \caption{Comparison of spatial regions of our TDA loss and existing losses}
  \vspace{-2pt}
  \label{fig:scales}
 \end{figure}

\section{Proposed method}
\label{sec:proposed}
We design a target driven adaptive (TDA) loss to enhance detection performance in local regions around targets and detection rates for targets with smaller scales and lower local contrasts. The proposed TDA loss consists of two key components: a patch-based mechanism and an adaptive adjusting strategy to the scales and local contrasts. As a patch-based mechanism, our proposed TDA loss is defined for each patch around the target, as shown in Fig. \ref{fig:scales}. We minimize the average of these losses over all targets in an image for training the model. It can be expressed as:
\begin{align}
    L_{T} &= \frac{1}{N}\sum_l^{N} L_{t},
\end{align}
where $N$ is the number of targets per image. The proposed TDA loss facilitates learning of local image context. By treating each target independently and equally, the TDA loss enhances detection performance for small targets even in multi-target scenarios. To implement the proposed TDA loss, we first perform segmentation on the label image using the spaghetti labeling method \cite{bolelli2019spaghetti}. We then compute bounding boxes for each target and dilate them by d pixels, where d is randomly chosen between 2 and 5 pixels for data augmentation. Using these dilated bounding boxes, we crop the image and ground-truth label, resize to a fixed size ($48 \times 48$), and then calculate the loss $L_{t}$.

The proposed TDA loss adaptively adjusts the loss value to emphasize small and low-contrast targets. 
The proposed TDA loss can be formulated as follows:
\begin{align}
    L_{t} &= - (1-I_{t}^{p_{t}})\cdot log(I_{t}),  \\
    I_{t} &= \frac{|A_{t, p}\cap A_{t, g}|}{|A_{t, p}\cup A_{t, g}|},\\
    p_{t} &= 1+\sigma(-\frac{s_{t}}{s_{mean}})+\sigma(-\frac{c_{t}}{c_{mean}}),
\end{align}
where $I_{t}$ is a soft IoU and $A_{t, p}$ and $A_{t, g}$ are the sets of predicted pixel values and ground-truth pixel values, respectively. $s_{t}$ is the size of each target and $s_{mean}$ is the mean size of all targets in the training dataset. We define scale as the number of pixels belonging to the target. $c_{t}$ is the local contrast value of each target and $c_{mean}$ is the mean local contrast of all targets in the training dataset. We define local contrast as the difference between the mean pixel value of the target and that of the background within the patch. $\sigma$ is a sigmoid function. The adaptive exponent $p_t$ is computed directly from each target’s scale and local contrast. The parameter $p_t$ smoothly increases for smaller or lower-contrast targets, consequently causing larger loss values and guiding the model to focus on those harder samples.

The proposed TDA loss and existing losses focus on different scales, as shown in Fig. \ref{fig:scales}. We combine the proposed TDA and the SLS loss $L_{S}$ to enhance the detection performance of the overall image and individual targets simultaneously. We use a weighted sum of our loss and the SLS loss for training. The total loss $L_{total}$ is expressed as:
\begin{align}
    L_{total} = L_{S} + w_T L_{T},
\end{align}
where $w_T$ is a weight factor for the proposed TDA loss. We examined $w_T$ values between 0.1 and 0.4 using the IRSTD-1k dataset, and found that 0.2 produced the highest IoU. Therefore, we adopted this value for $w_T$.

\section{Experiments}
\label{sec:experiments}

\begin{table}[t]
    \caption{Comparison of our adaptive weighting and fixed parameters on detection performance on the IRSTD-1k dataset. The proposed adaptive $p_t$ achieves the best performance.}
    
   \label{table:ablation_loss_function}
   \begin{center}
    \begin{tabular}{l|ccc}\hline

     Method       &    IoU$\uparrow$ &  $P_d$$\uparrow$  &    $F_a$$\downarrow$   \\ 
     \hline 
     $p_t=1$ (fixed)   &    69.10 &    93.47 &    \multicolumn{1}{r}{8.98} \\
     $p_t=2$ (fixed) &    69.23 &    93.81 &    \multicolumn{1}{r}{10.24} \\
     $p_t=3$ (fixed) &    69.02 &    94.15 &    \multicolumn{1}{r}{12.34} \\
     $Adaptive$ $p_t$ (ours)            &   \textbf{69.50}  &  \textbf{95.18} & \multicolumn{1}{r}{\textbf{8.73}}  \\

     \hline
  \end{tabular}
   \end{center}
  \end{table}

We evaluated the proposed TDA loss through four experiments. First, we analyzed the parameter $p_t$ of the proposed TDA loss by comparing three fixed values and an adaptive setting. Next, we evaluated the detection performance of TDA loss against existing loss functions on four datasets. We also evaluated the robustness across different scale ranges and local contrast conditions. Finally, we visualized qualitative differences between the proposed and baseline losses. To further assess the generality of the our loss across different network architectures, we additionally evaluated the performance of our loss using three different network architectures and the results are included in the supplementary material.

\subsection{Experimental settings}
We used four publicly available datasets: IRSTD-1k, SIRST v1, SIRST v2, and NUDT-SIRST \cite{zhang2022isnet, li2022dense, dai2021asymmetric, dai2023one}. Each dataset contains 1001, 417, 1024, and 1327 images, respectively. We followed the dataset split protocols in prior works \cite{zhang2022isnet, liu2024infrared, li2022dense}. 
 Specifically, IRSTD-1k, SIRST v1, SIRST v2 and NUDT-SIRST were divided into 800/201, 341/86, 768/256, and 663/664 images for training and testing, respectively.

We employed three metrics that are widely used in IRSTD studies: intersection over union (IoU), false-alarm rate ($F_a$), and probability of detection ($P_d$) \cite{zhang2022isnet, liu2024infrared, li2022dense}. IoU measures the overlap between predicted and ground truth segmentation. $F_a$ quantifies the ratio of falsely predicted pixels to all image pixels, assessing the precision of the method. $P_d$ represents the ratio of correctly detected targets to all targets, indicating the recall capability of the method. We set the threshold for all models to 0.5 to evaluate these three metrics. For comparison between our method and existing methods, we additionally evaluated receiver operating characteristics (ROC) curves to analyze the changing trend of $P_d$ under varying $F_a$.

\begin{table*}[t]
\centering
\footnotesize
\caption{We compared our proposed loss and existing losses with the metrics of of IoU(\%), $P_d (\%)$, and $F_a (10^{-6})$. The \textbf{best} and \underline{second} values are highlighted in bold and underline, respectively. The results show that the proposed TDA loss enhance detection performance for four datasets.}
\begin{tabular}{l|ccc|ccc|ccc|ccc}
\hline
Method & \multicolumn{3}{|c|}{IRSTD-1k} & \multicolumn{3}{|c|}{SIRST v1} & \multicolumn{3}{|c}{SIRST v2}& \multicolumn{3}{|c}{NUDT-SIRST} \\
\cline{2-13} 
 & IoU$\uparrow$ & $P_d$$\uparrow$ & $F_a$$\downarrow$ & IoU$\uparrow$ & $P_d$$\uparrow$ & $F_a$$\downarrow$ &
 IoU$\uparrow$ & $P_d$$\uparrow$ & $F_a$$\downarrow$ &IoU$\uparrow$ & $P_d$$\uparrow$ & $F_a$$\downarrow$ \\
\hline
BCE loss \cite{zhang2021review}          &63.32 & 90.72 & \multicolumn{1}{r|}{\textbf{6.37}} &      68.44 & \multicolumn{1}{r}{97.24} & \multicolumn{1}{r|}{22.88}    & 66.86 & 90.06 & \multicolumn{1}{r|}{11.26} & 72.41 & 98.09 & \multicolumn{1}{r}{\textbf{5.35}} \\
Focal loss \cite{lin2017focal}        & 65.40 & 90.72 & \multicolumn{1}{r|}{7.28} &     69.98 & \multicolumn{1}{r}{98.16} & \multicolumn{1}{r|}{21.64}    & 66.04 & 90.06 & \multicolumn{1}{r|}{ \underline{4.41}} & 73.46 & 97.78 & \multicolumn{1}{r}{6.73} \\
Tversky loss ($\alpha, \beta$) = (0.3, 0.7) \cite{salehi2017tversky}     &    66.30 & 92.43 & \multicolumn{1}{r|}{15.25}    & 67.69 &  \multicolumn{1}{r}{98.16} & \multicolumn{1}{r|}{20.22} & 55.14 &  92.71 & \multicolumn{1}{r|}{58.19}  & 67.56 & 95.34 & \multicolumn{1}{r}{32.81} \\
Tversky loss ($\alpha, \beta$) = (0.7, 0.3) \cite{salehi2017tversky}     &    62.62 & \underline{93.81} & \multicolumn{1}{r|}{24.36}    & 67.78 &  \multicolumn{1}{r}{\underline{99.08}} & \multicolumn{1}{r|}{23.77} & 60.20 &  \textbf{94.03} & \multicolumn{1}{r|}{27.41}   & 72.46 & 98.09 & \multicolumn{1}{r}{22.36} \\
IoU loss \cite{huang2019batching}        & 65.57 & 87.62 & \multicolumn{1}{r|}{\underline{6.83}} &         \underline{70.61} & \multicolumn{1}{r}{98.16} & \multicolumn{1}{r|}{29.27}    & 68.25 & 90.06 & \multicolumn{1}{r|}{11.80}  & 73.56 & 96.19 & \multicolumn{1}{r}{13.40} \\
Dice loss \cite{sudre2017generalised}        & 64.60 & 89.00 & \multicolumn{1}{r|}{8.50} &       69.15 & \multicolumn{1}{r}{99.08} & \multicolumn{1}{r|}{20.93}    & 64.35 & 89.40 & \multicolumn{1}{r|}{16.27}  & 72.54 & 97.67 & \multicolumn{1}{r}{9.03} \\
SLS loss \cite{liu2024infrared}       & \underline{67.81} & 92.43 & \multicolumn{1}{r|}{12.75} &        \underline{67.96} & \multicolumn{1}{r}{\underline{99.08}} & \multicolumn{1}{r|}{37.39}    & \underline{68.88} & 93.37 & \multicolumn{1}{r|}{6.43}  & 72.86 & 97.88 & \multicolumn{1}{r}{22.70} \\
SLS loss + TDA loss (Ours) & \textbf{69.50} & \textbf{95.18} & \multicolumn{1}{r|}{8.73} & \textbf{72.80} & \multicolumn{1}{r}{\textbf{100.00}} & \multicolumn{1}{r|}{\textbf{10.29}} & \textbf{70.26} & \textbf{94.03} & \multicolumn{1}{r|}{\textbf{2.80}}   & \textbf{78.01} & \textbf{98.41} & \multicolumn{1}{r}{17.05} \\
\hline
\end{tabular}
\label{tab:comparison}
\end{table*}

We used a state-of-the-art network, MSHNet, with our proposed loss for IRSTD \cite{liu2024infrared}. MSHNet has a simple multi-head structure based on U-Net and achieves better detection performance and faster estimation time. All input images are resized to $256 \times 256$. We followed the hyper-parameter settings for optimization used in the MSHNet paper, and applied them with different losses for fair comparison. We used AdaGrad as the optimizer, with batch size, initial learning rate and number of epochs set to 4, 0.05 and 400, respectively. We applied a series of data augmentation techniques. These include random horizontal flip, random scaling (within a $\pm50\%$ range), and random cropping to force the model to learn from partial information. We additionally applied brightness and contrast augmentation within a $\pm50\%$ range to enhance robustness to various brightness and contrast conditions.

We compared the proposed method against several existing losses for IRSTD and commonly used losses in semantic segmentation tasks. We used BCE loss, Focal loss, Tversky loss, IoU loss, Dice loss, and the recently proposed SLS loss as existing methods \cite{liu2024infrared}. We set the hyperparameter $\gamma$ to 2.0 for Focal loss, as proposed in \cite{lin2017focal}. We set the hyperparameters $(\alpha, \beta)$ of Tversky loss to (0.3, 0.7) or (0.7, 0.3) as different weight cases.

\begin{figure}[t]
  \begin{center}
   \includegraphics[width=0.95\linewidth]{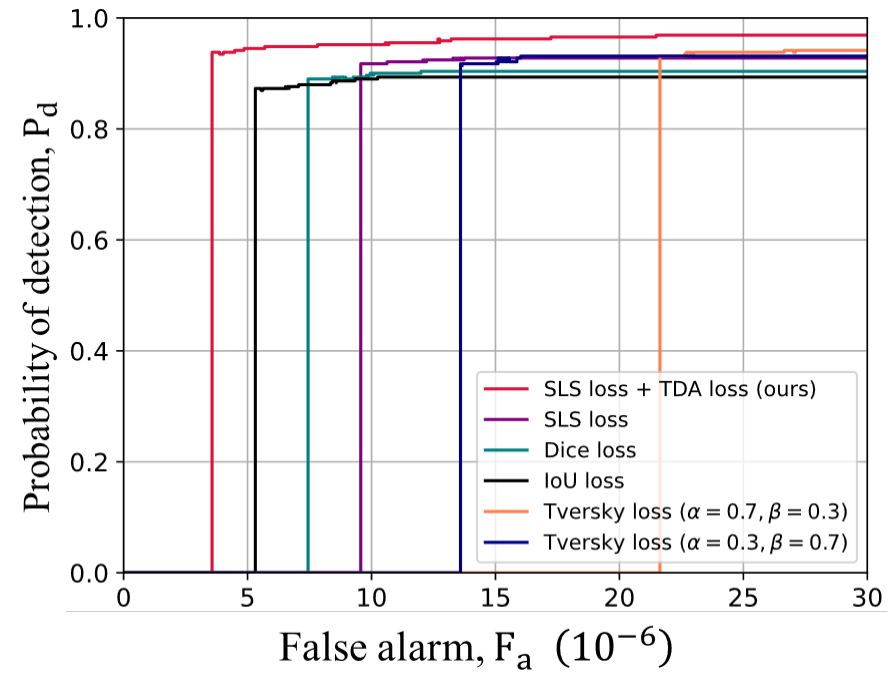}
  \end{center}
  \caption{ROC curve for the proposed and the prior global image level metric losses. We used IRSTD-1k dataset. The proposed TDA loss achieved better $P_d$ across various $F_a$.}
  \label{fig:roc}
  \vspace{-2pt}
 \end{figure}

\begin{figure}[t]
  \begin{center}
   \includegraphics[width=1.0\linewidth]{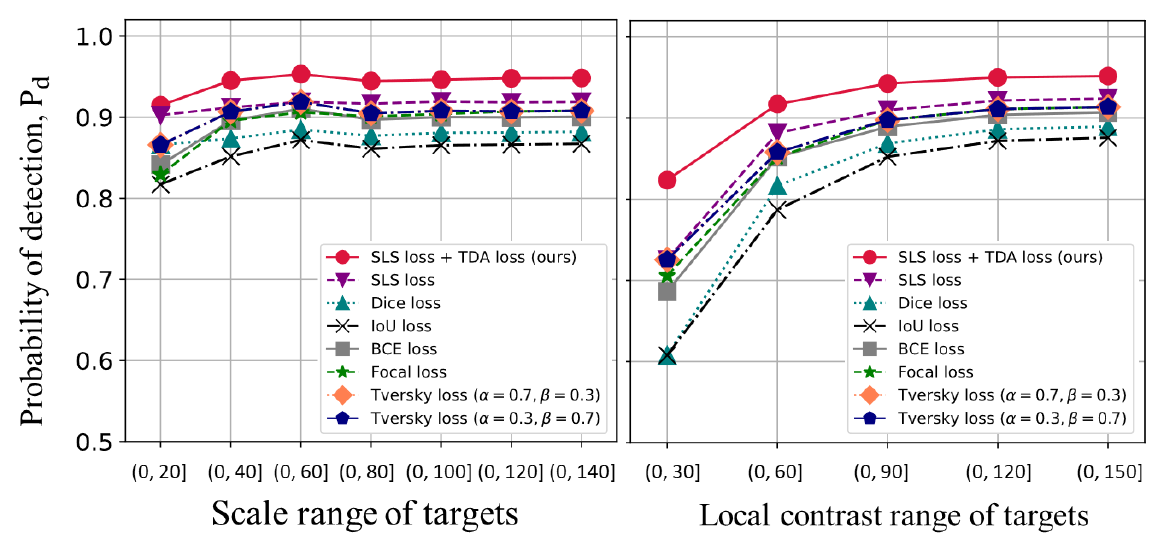}
  \end{center}
  \caption{Comparison of the proposed method and existing methods with $P_d$ at each scale range and local contrast range. The result shows our TDA loss enhance the detection performance for targets with smaller scale and lower local contrasts.}
  \label{fig:scale_contrast}
  \vspace{-2pt}
 \end{figure}
 
\subsection{Experimental results}

To quantify the improvement introduced by our adaptive strategy, we compared the detection performance between the adaptive and fixed $p_t$ settings. Fixed settings assign uniform emphasis across all targets, whereas our adaptive strategy increases the loss for smaller or lower-contrast targets based on Equation (4), encouraging the model to focus more on challenging examples. We evaluated three fixed values of $p_t$ : 1.0, 2.0, and 3.0. Table~\ref{table:ablation_loss_function} presents the results, showing that the adaptive setting achieves superior detection performance across IoU, $P_d$, and $F_a$.
These results indicate that our adaptive $p_t$ formulation guides the model better focus on challenging targets of small scale and low local contrast, thereby enhancing detection performance in IRSTD.

\begin{figure*}[t]
  \begin{center}
   \includegraphics[width=0.88\linewidth]{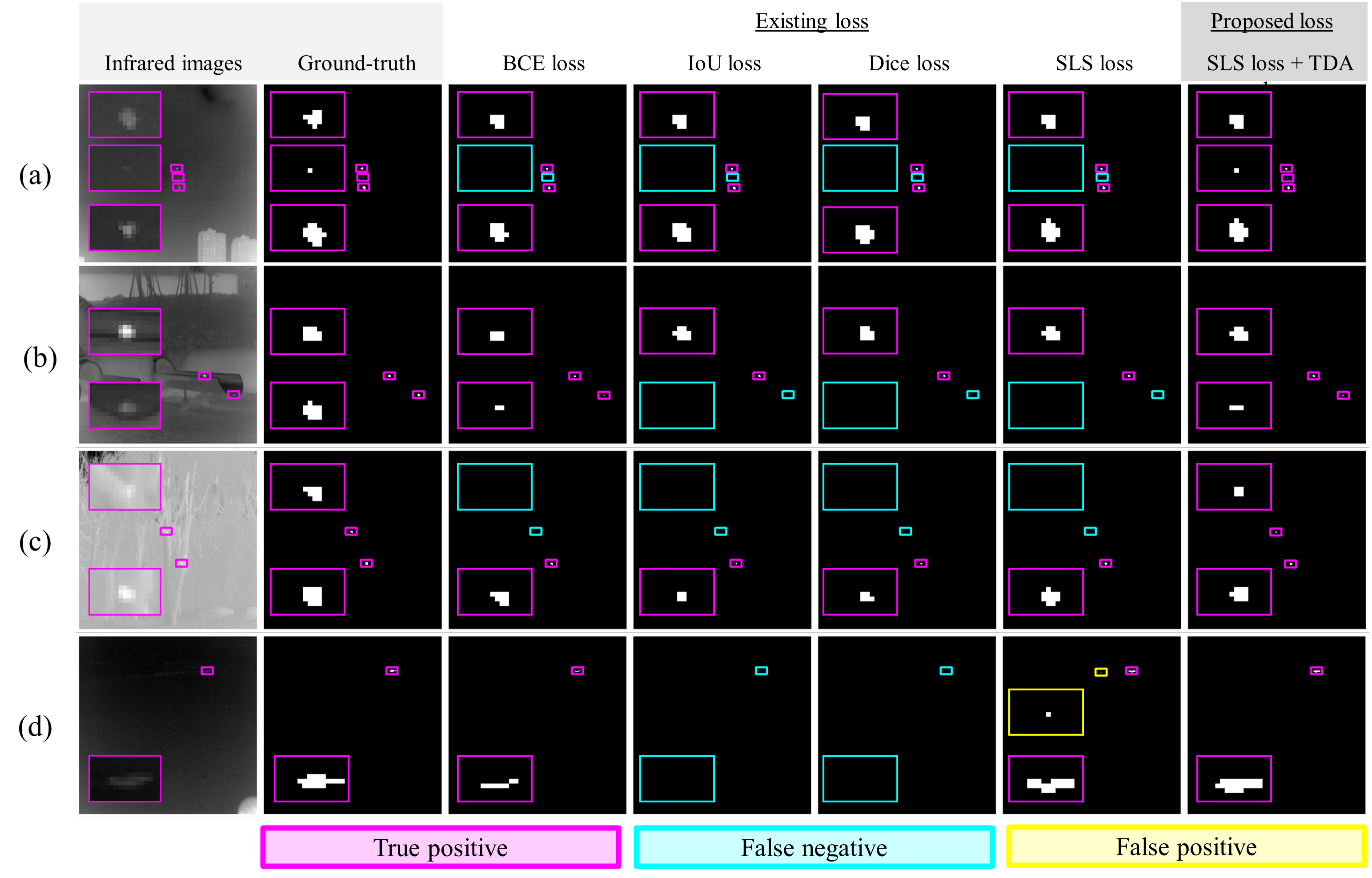}
  \end{center}
  \caption{Visual results of models trained by conventional losses for IRSTD-1k dataset. We can confirm that the TDA loss enhance the robustness to smalle scale and low local contrasts.}
  \label{fig:visual_comparison}
  \vspace{-2pt}
 \end{figure*}

Table \ref{tab:comparison} presents the comparison of the proposed TDA loss and existing losses with three metrics: IoU, $P_d$ and $F_a$. We used four datasets: IRSTD-1k, SIRST v1, SIRST v2, and NUDT-SIRST. Our proposed method achieved superior detection performance in terms of IoU and $P_d$. These results indicate that the proposed method enables the model to extract target shapes more accurately and improves detection performance for more challenging targets. In addition, to evaluate the generalization ability of the proposed TDA loss, we applied it to three additional network architectures: U-Net, UIU-Net, and SCTransNet \cite{wu2022uiu, yuan2024sctransnet, ronneberger2015u}. The results demonstrate that TDA loss consistently enhances detection performance across all architectures. Detailed results are provided in the supplementary material.

We compared the proposed TDA loss and existing global image level metric losses using ROC curve to evaluate comprehensive detection performance across various thresholds. Here, we do not compare the BCE loss and Focal loss because the ROC curves with these losses exhibited severe oscillations, making quantitative comparisons difficult. Fig. \ref{fig:roc} presents the comparative results. The results show that the proposed method achieved better detection performance across various false alarm rates compared to other existing losses. The proposed method also attained the highest probability of detection and lowest false alarm rate by appropriately selecting the threshold. These results suggest that the proposed method facilitates learning features that effectively discriminate between target and background regions.

To evaluate the robustness against small-scale and low-local-contrast targets, we assessed the probability of detection ($P_d$) across various scale ranges and local contrast ranges. Here, scale is defined as the number of pixels in a target, and local contrast is defined as the difference between the target’s mean intensity and that of the surrounding background measured within a bounding box dilated by three pixels. We compared the $P_d$ of each method for targets within the scale ranges of (0, 20), (0, 40), (0, 60), (0, 80), (0, 100), (0, 120), and (0, 140) pixels. The left side of Fig. \ref{fig:scale_contrast} shows that the proposed loss achieves the highest $P_d$ in every bin, confirming its superiority for small targets. We also evaluated the $P_d$ for targets within the local contrast ranges of (0, 30), (0, 60), (0, 90), (0, 120), and (0, 150). Fig.\ref{fig:scale_contrast} (right) again places our method on top across all ranges, demonstrating superior detection of low-contrast targets. These results demonstrate that our TDA loss effectively emphasize the smaller and lower-contrast targets and improve the performance for them while maintaining or even improving the performance for easier targets.

To qualitatively evaluate the proposed method, we visualized the detection results of the proposed approach and several representative comparative methods. As shown in Fig. \ref{fig:visual_comparison}, models trained with existing losses tend to fail to detect small scale and low local contrast targets in the images. The proposed TDA loss treats each target independently and equally, leading to scale-invariant detection. As shown in Fig. \ref{fig:visual_comparison} (a), the model trained with the TDA loss can detect smaller targets in multi-target scenarios. Furthermore, the proposed TDA loss focuses on challenging targets with low local contrast by adaptively adjusting the loss value and improves the detection performance for these targets. Consequently, the model trained with our loss can detect targets with low local contrast, as shown in Fig. \ref{fig:visual_comparison} (b), (c), and (d).

\section{Conclusion}
\label{sec:conclusion}
In this study, we introduced the TDA loss for infrared small target detection, designed to enhance detection performance, particularly for challenging small scale and low local contrast targets. The proposed TDA loss comprises two key components: a patch-based feature learning mechanism, and a dynamic loss adaptation strategy to individual target characteristics. The first component leads the model to focus on local regions around targets and the second component improves detection performance for targets with small scale and low local contrast. We integrated the proposed TDA loss with an existing SLS loss to train a deep learning model. The experimental results demonstrated significant improvements in detection performance, especially for challenging targets.

\vfill\pagebreak

\bibliographystyle{IEEEbib}
\bibliography{refs_short}

\begin{thebibliography}{10}

\bibitem{zhao2022single}
Mingjing Zhao, Wei Li, Lu~Li, Jin Hu, Pengge Ma, and Ran Tao,
\newblock ``Single-frame infrared small-target detection: A survey,''
\newblock {\em IEEE TGRS}, vol. 10, no. 2, pp. 87--119, 2022.

\bibitem{zhang2021review}
Ke~Zhang, Shuyan Ni, Dashuang Yan, and Aidi Zhang,
\newblock ``Review of dim small target detection algorithms in single-frame
  infrared images,''
\newblock in {\em IEEE IMCEC}. IEEE, 2021, vol.~4, pp. 2115--2120.

\bibitem{yu2022pay}
Chuang Yu, Yunpeng Liu, Shuhang Wu, Xin Xia, Zhuhua Hu, Deyan Lan, and Xin Liu,
\newblock ``Pay attention to local contrast learning networks for infrared
  small target detection,''
\newblock {\em IEEE GRSL}, vol. 19, pp. 1--5, 2022.

\bibitem{dai2021attentional}
Yimian Dai, Yiquan Wu, Fei Zhou, and Kobus Barnard,
\newblock ``Attentional local contrast networks for infrared small target
  detection,''
\newblock {\em IEEE TGRS}, vol. 59, no. 11, pp. 9813--9824, 2021.

\bibitem{hou2021ristdnet}
Qingyu Hou, Zhipeng Wang, Fanjiao Tan, Ye~Zhao, Haoliang Zheng, and Wei Zhang,
\newblock ``{RISTDnet}: Robust infrared small target detection network,''
\newblock {\em IEEE GRSL}, vol. 19, pp. 1--5, 2021.

\bibitem{han2019local}
Jinhui Han, Saed Moradi, Iman Faramarzi, Chengyin Liu, Honghui Zhang, and Qian
  Zhao,
\newblock ``A local contrast method for infrared small-target detection
  utilizing a tri-layer window,''
\newblock {\em IEEE GRSL}, vol. 17, no. 10, pp. 1822--1826, 2019.

\bibitem{dai2017reweighted}
Yimian Dai and Yiquan Wu,
\newblock ``Reweighted infrared patch-tensor model with both nonlocal and local
  priors for single-frame small target detection,''
\newblock {\em IEEE J-STARS}, vol. 10, no. 8, pp. 3752--3767, 2017.

\bibitem{zhang2022isnet}
Mingjin Zhang, Rui Zhang, Yuxiang Yang, Haichen Bai, Jing Zhang, and Jie Guo,
\newblock ``{ISNet}: Shape matters for infrared small target detection,''
\newblock in {\em CVPR}, 2022, pp. 877--886.

\bibitem{liu2024infrared}
Qiankun Liu, Rui Liu, Bolun Zheng, Hongkui Wang, and Ying Fu,
\newblock ``Infrared small target detection with scale and location
  sensitivity,''
\newblock in {\em CVPR}, 2024, pp. 17490--17499.

\bibitem{li2022dense}
Boyang Li, Chao Xiao, Longguang Wang, Yingqian Wang, Zaiping Lin, Miao Li, Wei
  An, and Yulan Guo,
\newblock ``Dense nested attention network for infrared small target
  detection,''
\newblock {\em IEEE TIP}, vol. 32, pp. 1745--1758, 2022.

\bibitem{wu2022uiu}
Xin Wu, Danfeng Hong, and Jocelyn Chanussot,
\newblock ``{UIU-Net}: {U-Net} in {U-Net} for infrared small object
  detection,''
\newblock {\em IEEE TIP}, vol. 32, pp. 364--376, 2022.

\bibitem{wang2019miss}
Huan Wang, Luping Zhou, and Lei Wang,
\newblock ``Miss detection vs. false alarm: Adversarial learning for small
  object segmentation in infrared images,''
\newblock in {\em ICCV}, 2019, pp. 8509--8518.

\bibitem{yuan2024sctransnet}
Shuai Yuan, Hanlin Qin, Xiang Yan, Naveed Akhtar, and Ajmal Mian,
\newblock ``{SCTransNet}: Spatial-channel cross transformer network for
  infrared small target detection,''
\newblock {\em IEEE TRGS}, 2024.

\bibitem{dai2021asymmetric}
Yimian Dai, Yiquan Wu, Fei Zhou, and Kobus Barnard,
\newblock ``Asymmetric contextual modulation for infrared small target
  detection,''
\newblock in {\em CVPRW}, 2021, pp. 950--959.

\bibitem{arpit2017closer}
Devansh Arpit et~al.,
\newblock ``A closer look at memorization in deep networks,''
\newblock in {\em ICML}, 2017, pp. 233--242.

\bibitem{wang2022iaanet}
Kewei Wang, Shuaiyuan Du, Chengxin Liu, and Zhiguo Cao,
\newblock ``{Interior} {Attention-Aware} {Network} for {Infrared} {Small}
  {Target} {Detection},''
\newblock {\em IEEE TGRS}, vol. 60, pp. 1--13, 2022.

\bibitem{dai2023one}
Yimian Dai, Xiang Li, Fei Zhou, Yulei Qian, Yaohong Chen, and Jian Yang,
\newblock ``One-stage cascade refinement networks for infrared small target
  detection,''
\newblock {\em IEEE TGRS}, vol. 61, pp. 1--17, 2023.

\bibitem{huang2019batching}
Yifeng Huang, Zhirong Tang, Dan Chen, Kaixiong Su, and Chengbin Chen,
\newblock ``Batching soft {IoU} for training semantic segmentation networks,''
\newblock {\em IEEE Signal Process. Lett.}, vol. 27, pp. 66--70, 2019.

\bibitem{sudre2017generalised}
Carole~H Sudre, Wenqi Li, Tom Vercauteren, Sebastien Ourselin, and
  M~Jorge~Cardoso,
\newblock ``Generalised dice overlap as a deep learning loss function for
  highly unbalanced segmentations,''
\newblock in {\em DLMIA-MLCDS}. Springer, 2017, pp. 240--248.

\bibitem{lin2017focal}
Tsung-Yi Lin, Priya Goyal, Ross Girshick, Kaiming He, and Piotr Dollar,
\newblock ``{Focal Loss} for dense object detection,''
\newblock {\em IEEE TPAMI}, vol. 42, no. 2, pp. 318--327, 2020.

\bibitem{jeune2023rethinking}
Pierre~Le Jeune and Anissa Mokraoui,
\newblock ``Rethinking intersection over union for small object detection in
  few-shot regime,''
\newblock {\em arXiv preprint arXiv:2307.09562}, 2023.

\bibitem{bolelli2019spaghetti}
Federico Bolelli, Stefano Allegretti, Lorenzo Baraldi, and Costantino Grana,
\newblock ``Spaghetti labeling: Directed a cyclic graphs for block-based
  connected components labeling,''
\newblock {\em IEEE TIP}, vol. 29, pp. 1999--2012, 2020.

\bibitem{salehi2017tversky}
Seyed Sadegh~Mohseni Salehi, Deniz Erdogmus, and Ali Gholipour,
\newblock ``Tversky loss function for image segmentation using {3D} fully
  convolutional deep networks,''
\newblock in {\em MLMI}. Springer, 2017, pp. 379--387.

\bibitem{ronneberger2015u}
Olaf Ronneberger, Philipp Fischer, and Thomas Brox,
\newblock ``{U-Net}: Convolutional networks for biomedical image
  segmentation,''
\newblock in {\em MICCAI}. Springer, 2015, pp. 234--241.

\end{thebibliography}


\maketitlesupplementary 

\vspace{1em}  


\noindent
\begin{minipage}{\textwidth}
    \centering
    \captionof{table}{We compared our proposed loss and existing losses with the metrics of IoU ($\%$), $P_d (\%)$, and $F_a (10^{-6})$. The best and second values are highlighted in bold and underline, respectively.}
    %
    \label{tab:comp_archi}
    \vspace{0.5em}  
    \begin{tabular}{l|ccc|ccc|ccc}
      \hline
        Method & \multicolumn{3}{|c|}{U-Net \cite{ronneberger2015u}} & \multicolumn{3}{|c|}{UIUNet \cite{wu2022uiu}} & \multicolumn{3}{|c}{SCTransNet \cite{yuan2024sctransnet}} \\
        \cline{2-10} 
         & IoU$\uparrow$ & $P_d$$\uparrow$ & $F_a$$\downarrow$ & IoU$\uparrow$ & $P_d$$\uparrow$ & $F_a$$\downarrow$ & IoU$\uparrow$ & $P_d$$\uparrow$ & $F_a$$\downarrow$ \\
        \hline
        BCE loss \cite{zhang2021review}        & 64.76 & 88.65 & \textbf{10.32} & 64.56 & 88.77 & 10.38 & 67.03 & 89.00 & \textbf{6.98} \\
        IoU loss \cite{huang2019batching}      & 63.14 & 91.40 & \underline{19.58} & \underline{63.70} & 88.43 & 22.88 & 67.38 & 88.60 & 18.97 \\
        Dice loss \cite{sudre2017generalised}  & 62.19 & 91.75 & 12.37 & 62.27 & 90.81 & 22.64 & 67.72 & \underline{89.69} &  16.64\\
        SLS loss \cite{liu2024infrared}       & \underline{66.24} & 92.43 & 38.10 & \underline{65.20} & \underline{90.81} & 25.02 & \underline{68.22} & 88.65& 21.18 \\
        SLS loss + TDA loss (Ours)    & \textbf{69.48} & \textbf{94.84} & 14.04 & \textbf{66.96} & \textbf{92.17} & \textbf{9.56} & \textbf{68.59} & \textbf{91.75} & \underline{9.41} \\
        \hline
    \end{tabular}
\end{minipage}

\vspace{1em}


The proposed TDA loss can be applied to prior segmentation models developed for IRSTD and enhance their detection performance. 
To validate this claim, we evaluated TDA loss on three other architectures: a simple U-Net, and two state-of-the-art models, UIU-Net and SCTransNet \cite{wu2022uiu, yuan2024sctransnet, ronneberger2015u}. U-Net has a simple encoder and decoder architecture, and it is able to capture both global context and fine-grained features. UIUNet consists of a U-Net embedded with smaller U-Nets. UIUNet effectively extracts multi-scale information using a nested architecture. SCTransNet introduces spatial-channel cross transformer blocks. SCTransNet encodes global context information and improves detection performance for targets with high similarity to the background. 


We used Adam as the optimizer for all models. For U-Net, we adopted the settings from the open-source implementation\footnote{\url{https://github.com/XinyiYing/BasicIRSTD}}: learning rate 0.0005, batch size 16, 400 epochs, with the learning rate halved at epochs 200 and 300. For UIUNet and SCTransNet, we followed their original papers, using a learning rate of 0.001. UIUNet used batch size 3 and 500 epochs; SCTransNet used batch size 16 and 1000 epochs, with cosine annealing to 0.00001.

Table \ref{tab:comp_archi} shows the detection performance of our TDA loss and existing losses with three different architectures: UNet, UIUNet, and SCTransNet. We used IRSTD-1k dataset for training and evaluation. The results show that our TDA loss enhances the detection performance for three different networks. These results indicate that the proposed method enables the model to extract target shapes more accurately and improves detection performance for more challenging targets.


To verify the applicability of our TDA loss to existing loss functions, we evaluated its performance in combination with IoU loss, Dice loss, and BCE loss. We used MSHNet as a backbone model. Table \ref{table:loss_combination} compares the detection performance with and without TDA loss on the IRSTD-1k dataset. The result demonstrates that our TDA loss comprehensively improves performance and can be effectively integrated with various existing losses for IRSTD. 

\begin{table}[h]
    \caption{  
    Comparison of detection performance on IRSTD-1k with and without the proposed TDA loss. The proposed TDA loss consistently improves performance when combined with BCE, IoU, and Dice losses. The better values for each metric are highlighted in bold.}
    
   \label{table:loss_combination}
   \begin{center}
   \footnotesize
    \begin{tabular}{l|ccc}\hline
     Method       &    IoU$\uparrow$ &  $P_d$$\uparrow$  &    $F_a$$\downarrow$   \\ 
     \hline 
     BCE loss             &    63.32 &    90.72 &    \textbf{6.37} \\
     BCE loss + TDA loss  &    \textbf{65.13} (+1.81) &   \textbf{96.21} (+5.49)&    18.52 (+12.15) \\
     \hline 
     IoU loss  &    65.57 &    87.62 &    \textbf{6.83} \\
     IoU loss + TDA loss     &    \textbf{67.31} (+1.74)&    \textbf{91.06} (+3.44)&    
     9.71 (+2.88) \\
     \hline 
     Dice loss           &    64.60 &    89.00 &    8.50 \\
     DIce loss + TDA loss            &    \textbf{67.34} (+2.74) &  \textbf{90.72} (+1.72)& \textbf{7.43} (-1.07)  \\

     \hline
  \end{tabular}
   \end{center}
  \end{table}

\end{document}